%% file: m1488_arxiv_revised.tex
\documentclass[]{ecai}
\usepackage[T1]{fontenc}
\usepackage{graphicx}
\def\Args{Args}
\def\Att{Att}

\def\wrt{wrt}

\newcommand{\abaf}{\ensuremath{\langle {\cal L}, \, {\cal R}, \, {\cal A},\, \overline{ \vrule height 5pt depth 3.5pt width 0pt \hskip0.5em\kern0.4em}\rangle}}
\def\contrary{\overline{ \vrule height 5pt depth 3.5pt width 0pt \hskip0.5em\kern0.4em}}

\newcommand{\abafpone}{\ensuremath{\langle {\cal L}, \, {\cal R}', \, {\cal A},\, \overline{ \vrule height 5pt depth 3.5pt width 0pt \hskip0.5em\kern0.4em}\rangle}}

\newcommand{\abafp}{\ensuremath{\langle {\cal L}', \, {\cal R}', \, {\cal A}',\, \overline{ \vrule height 5pt depth 3.5pt width 0pt \hskip0.5em\kern0.4em}'\rangle}}

\newcommand{\argur}[3]{#1 \vdash_{#3} #2} 
\newcommand{\argu}[2]{#1 \vdash #2} 
\newcommand{\ruleset}{\ensuremath{R}}
\newcommand{\sent}{\ensuremath{s}}

\newcommand{\asm}{\ensuremath{a}}
\newcommand{\asmset}{\ensuremath{A}}

\newcommand{\pif}{:\hspace{-2.5pt}\text{--}}

\newcommand{\Babalearn}{\textit{ASP-ABAlearn}_{B}}
\newcommand{\EBabalearn}{\textit{ASP-ABAlearn}_{\textit{BE}}}

\usepackage[ruled,resetcount,linesnumbered]{algorithm2e}
\let\oldnl\nl
\newcommand{\nonl}{\renewcommand{\nl}{\let\nl\oldnl}}
\usepackage{amsmath}
\usepackage{amsthm}
\usepackage{amssymb}
\usepackage{graphicx}
\usepackage[colorlinks=true, allcolors=blue]{hyperref}
\usepackage{xcolor}
\usepackage{cancel}
\usepackage{todonotes}
\usepackage{enumitem}
\usepackage{booktabs}

\newcommand{\A}{\mathcal{A}}
\newcommand{\T}{\mathcal{T}}
\newcommand{\LL}{\mathcal{L}}
\newcommand{\RR}{\mathcal{R}}

\newcommand{\sABA}{\langle\RR,\A,\contrary\rangle}
\newcommand{\sABAp}{\langle \RR',\A',\contrary'\rangle}
\newcommand{\sABAo}{\langle \RR_0,\A_0,\contrary^{_0}\rangle}

\newcommand{\IF}{\ensuremath{\leftarrow}}

\newcommand{\vdashacc}{\ensuremath{\models}}

\newcommand{\RRl}{\RR_{l}}
\newcommand{\RRt}{\RR_{t}}
\newcommand{\RRin}{\RR_{0}}

\newcommand{\Ep}{\mathcal{E}^+}
\newcommand{\En}{\mathcal{E}^-}
\newcommand{\EE}{\langle\Ep,\En\rangle}

\newtheorem{theorem}{Theorem}
\newtheorem{example}{Example}
\newtheorem{definition}{Definition}

\newtheorem{proposition}{Proposition}

\makeatletter
\patchcmd{\@algocf@start}
  {-1.5em}
  {-0.1em}
  {}{}
\makeatother

\begin{document}

\begin{frontmatter}




\title{Learning Brave Assumption-Based Argumentation Frameworks via ASP}


\author[A]{\fnms{Emanuele}~\snm{De Angelis}\orcid{0000-0002-7319-8439}\thanks{Corresponding Author. Email: emanuele.deangelis@iasi.cnr.it}}

\author[A]{\fnms{Maurizio}~\snm{Proietti}\orcid{0000-0003-3835-4931}}
\author[B]{\fnms{Francesca}~\snm{Toni}\orcid{0000-0001-8194-1459}}

\address[A]{CNR-IASI, Rome, Italy}
\address[B]{Imperial, London, UK}


\begin{abstract}
Assumption-based Argumentation (ABA)  is advocated as a unifying formalism for various forms of non-monotonic reasoning, including logic programming. 
It allows capturing defeasible knowledge, subject to argumentative debate. While, in much existing work,
ABA frameworks are given up-front, 
in this paper we focus on the problem of automating their learning 
from background knowledge and positive/negative examples. 
Unlike prior work, 
we newly frame the problem in terms of brave reasoning under stable extensions for ABA.
We present a novel algorithm based on transformation rules (such as Rote Learning, Folding, Assumption Introduction and Fact Subsumption) and 
an implementation thereof that makes use of 
Answer Set Programming.
Finally, we compare our technique to state-of-the-art ILP systems that learn 
defeasible knowledge.

\end{abstract}

\end{frontmatter}

\section{Introduction}

Assumption-based Argumentation (ABA)~\cite{ABA,ABAhandbook,ABAbook,ABAtutorial} is a form of structured argumentation broadly advocated as a unifying formalism for various formalisations of non-monotonic reasoning, including logic programming~\cite{ABA}. It allows capturing defeasible knowledge subject to argumentative debate, whereby arguments are deductions built from rules and supported by assumptions and, in order to be ``accepted'', they need to deal with attacks from other arguments (for the contraries of assumptions in their support). 

In much existing work,
fully-formed 
ABA frameworks are given up-front, e.g. to model medical guidelines~\cite{guidelines} or planning~\cite{fan-planning}.
Instead, in this paper we
focus on the problem of automating their learning from background
knowledge and positive and negative examples.
Specifically, we consider the 
recent formulation of \emph{ABA Learning}~\cite{MauFraILP22} 
for learning 
ABA frameworks from a background knowledge, in the form of an initial ABA framework, and positive and negative examples, in the form of sentences in the language of the background knowledge. The goal of ABA Learning is to build a larger ABA framework than the background knowledge from which  arguments for all positive examples can be ``accepted''  and no arguments for any of the negative examples can be ``accepted''. In this paper, for a specific form of ABA frameworks corresponding to logic programs~\cite{ABA}, we focus on a specific form of ``acceptance'', given by brave (or credulous, as commonly referred to in the argumentation literature) reasoning under the argumentation semantics of stable extensions~\cite{ABA,ABAhandbook}. 

We base our approach to brave ABA Learning on transformation rules, in the spirit of~\cite{MauFraILP22}.
We leverage on the well known correspondence~\cite{ABA} between stable extensions in the logic programming instance of ABA and Answer Set Programs (ASP)~\cite{GeL91} to outline a novel implementation strategy for the form of ABA Learning we consider, pointing out along the way restrictions on ABA Learning enabling the use of 
ASP. We also show experimentally, on some standard benchmarks, that the resulting $\Babalearn$ system performs well in comparison with ILASP~\cite{LawRB14}, a state-of-the-art system  in inductive logic programming (ILP) able to learn non-stratified logic programs.
In summary, our main contributions are:
(i) a novel definition of \emph{brave ABA Learning};
(ii) a novel (sound and terminating) $\Babalearn$ system for carrying out brave ABA Learning in ASP;
(iii) an empirical evaluation of $\Babalearn$ showing its strengths in comparison with the ILASP system.
All proofs are given in the Appendix.
The $\Babalearn$ system is available at \cite{ecai2024codedata}.

\section{Related Work}
Forms of ABA Learning have already been considered in~\cite{DPT23a,MauFraILP22,TPT23}. Like \cite{MauFraILP22} we rely upon transformation rules, adopting a variant of Subsumption and omitting to use Equality Elimination. However, we adopt a novel formulation of \emph{brave} ABA Learning. Like \cite{DPT23a} we use ASP as the basis for implementing ABA Learning, but, again, we focus on brave, rather than cautious, ABA Learning.  Finally,  \cite{TPT23} focuses on cautious ABA Learning and uses Python rather than ASP
.

 Our strategy for ABA Learning differs from other works learning argumentation frameworks, e.g. \cite{dear,DimopoulosK95,
 Tony-ArgML22}, 
 in that it learns a different type of argumentation frameworks and, 
 also, is based on brave 
 reasoning rather than cautious (a.k.a. sceptical). Also, these approaches do not make use  of ASP for supporting learning  algorithmically. 

ABA can be seen as performing abductive reasoning (as assumptions are hypotheses open for debate).  
Other approaches combine  
abductive  and inductive learning~\cite{Ray09}, but they do not learn ABA frameworks.  
Moreover, while using a definition of abduction wrt brave/credulous stable model semantics, \cite{Ray09} does not identify any 
property of brave induction and focuses on case studies, in the context of the event calculus, with a unique answer set (where brave and cautious reasoning coincide).
Some other approaches learn abductive logic programs~\cite{InoueH00}, which rely upon assumptions, like ABA
. 
A~formal comparison with these methods is left for future work.

ABA captures several non-monotonic reasoning formalisms, thus ABA Learning is related to other methods for learning non-monotonic formalisms. Some of these methods, e.g. \cite{InoueK97,Sakama00}, do not make use of ASP, and others, e.g. \cite{ShakerinSG17}, learn stratified logic programs with a unique stable model. Some others, 
e.g. \cite{LawRB14,Sakama05,SakamaI09}, do consider ASP programs with multiple stable models. 
Amongst these approaches, ILASP \cite{LawRB14} can be tailored to perform both brave and cautious induction of ASP programs, whereas
\cite{Sakama05} performs cautious induction in ASP, 
and 
the approach of \cite{SakamaI09} can perform brave induction in ASP
.
Induction from answer sets is also considered by Otero \cite{Otero01}.
In this paper, the use of ASP is mainly aimed at implementing some specific tasks of our ABA Learning strategy
(e.g. its use of the Rote Learning and Assumption Introduction transformation rules). 
More in general, differently from other learning approaches in ASP, our learning strategy is based on argumentative reasoning.
A formal and empirical comparison with these methods  is left for future work.

\section{Background}
\label{sec:background}

\subsection{Answer Set Programs}
We use 
ASP~\cite{GeL91
} consisting of rules of the form

\smallskip
$\mathtt{p ~\texttt{:-} 
~q_1,\dots, q_k, ~not~q_{k+1},\ldots, ~not~q_{n} 
}$ \quad \quad or 

$\mathtt{  \texttt{:-} 
~q_1,\dots, q_k, ~not~q_{k+1},\ldots, ~not~q_{
n} 
}$

\smallskip
\noindent
where ${\tt p }$, ${\tt q}_1$, $\dots$, ${\tt q_n}$,
are atoms, 
$\mathtt{k}\geq 0$,
$\mathtt{n}\geq 0$,
and ${\tt not }$ denotes negation as failure.
We assume that the reader is familiar with the \emph{stable model semantics} for ASP
~\cite{GeL91}, and we call \textit{answer set} of $P$ any set of ground atoms assigned to $P$ by that semantics.
$P$ is said to be \emph{satisfiable}, denoted $\textit{sat}(P)$, if it has an answer set, and  \emph{unsatisfiable} otherwise.
An atom ${\tt p }$ is a \textit{brave consequence} of $P$ if there exists an answer set $A$ of $P$ such that ${\tt p } \in A$.

\subsection{Assumption-Based Argumentation (ABA) 
}
%
An {\em ABA framework} (as originally proposed in \cite{ABA}, but presented here following 
\cite{ABAbook,ABAtutorial} and 
\cite{ABAhandbook}) is a tuple \abaf{}
such that

\vspace*{-0.2cm}
\begin{itemize}
\item 
$\langle \LL, \RR\rangle$ is a deductive system,
 where $\LL$ is a \emph{language} and $\RR$ is a set of
 \emph{(inference) rules} of the form $\sent_0 \leftarrow \sent_1,\ldots, \sent_m $ ($m \ge 0, \sent_i \in \LL$, for $1\leq i \leq m$); 

\item 
$\A$ $\subseteq$ $\LL$ is a (non-empty) 
set
of {\em assumptions};\footnote{The non-emptiness requirement
 can always be satisfied by including in $\A$ a \emph{bogus assumption}, with its own contrary, neither occurring elsewhere
 ~\cite{ABAtutorial}. 
} 

\item 
$\contrary$ is a \textit{total mapping} from $\A$ into
 $\LL$, where $\overline{\asm}$ is the {\em contrary} of $\asm$, for $\asm
  \in \A$ (also denoted as $\{a \mapsto \overline{a} \mid a \in \A\}$).
\end{itemize}
\vspace*{-0.2cm}
Given a rule $\sent_0 \gets \sent_1, \ldots,
\sent_m$, $\sent_0$ is 
the {\em head} 
and $\sent_1,\ldots, \sent_m$ 
is 
the {\em body}; 
if $m=0$ then the  body is said to be {\em empty} (represented as $\sent_0 \gets$ or  $\sent_0 \gets true$) and the rule is called a \emph{fact}.
In this paper we focus on \emph{flat} ABA frameworks, where assumptions are not heads  of rules.  
Elements of $\LL$ can be any sentences, but in this paper we  focus on 
ABA frameworks where $\LL$ is a finite set of ground atoms. 
However, in the spirit of logic programming, we will use \emph{schemata} for rules, assumptions and contraries, using variables to represent compactly all instances over some underlying universe. By $\textit{vars}(E)$ we denote the set of variables occurring in atom, rule, or rule body~$E$.




\begin{example}\label{ex:NixonBK} 
We consider a variant of the well-known Nixon diamond problem~\cite{nixon}, formalised as the 
ABA framework with

$\LL =\{$$\textit{quaker}(X), \textit{democrat}(X), \textit{republican}(X), \textit{person}(X), $

\phantom{$\LL =\{$}$\textit{votes}\_\textit{dem}(X), \textit{votes}\_\textit{rep}(X), \textit{normal}\_\textit{quaker}(X)$ 

\phantom{$\LL =\{$}$\mid X\in\{a,b,c,d,e\}\}$

$\RR = \{ 
\rho_1: \textit{quaker}(a)\leftarrow,\,
\rho_2: \textit{quaker}(b)\leftarrow,\,
\rho_3: \textit{quaker}(e)\leftarrow,$

\phantom{$\RR = \{$}$\rho_4: \textit{democrat}(c)\leftarrow,\quad \rho_5: \textit{republican}(a)\leftarrow,$

\phantom{$\RR = \{$}$\rho_6: \textit{republican}(b)\leftarrow,\quad \rho_7: \textit{republican}(d)\leftarrow,$

\phantom{$\RR = \{$}$\rho_{8}: \textit{democrat}(X) \leftarrow \textit{person}(X),\ \textit{votes\_dem}(X),$

\phantom{$\RR = \{$}$\rho_{9}: \textit{republican}(X) \leftarrow \textit{person}(X),\ \textit{votes}\_\textit{rep}(X),$

\phantom{$\RR = \{$}$\rho_{10}: \textit{pacifist}(X) \leftarrow  \textit{quaker}(X), \textit{normal\_quaker(X)}$

\phantom{$\RR = \{$}$\rho_{11}: \textit{person}(X)\leftarrow \ \ \ \mid X \in \{a,b,c,d,e\}\}$

$\A = \{\textit{votes\_dem}(X), \textit{votes\_rep}(X), \textit{normal\_quaker(X)}$ 

\phantom{$\A = \{$}$ \mid X \in \{a,b,c,d,e\}\}$

\vspace{1mm}
$\overline{\textit{votes}\_\textit{dem}(X)} \!=\! \textit{republican}(X),$

$\overline{\textit{votes}\_\textit{rep}(X)}\!=\! \textit{democrat}(X),\quad$

$ \overline{\textit{normal\_quaker(X)}} = \textit{abnormal\_quaker}(X) \mid X \in \{a,b,c,d,e\}$.
\end{example}

The semantics of flat ABA frameworks is given 
by ``acceptable'' extensions, i.e.~sets of \emph{arguments} able to ``defend'' themselves against {\em attacks}, in some sense, as determined by the chosen semantics.
Intuitively, arguments are deductions of claims using rules and 
supported by assumptions, and  attacks are directed at the
assumptions in the support of arguments.  
More formally, following
\cite{ABAhandbook,ABAbook,ABAtutorial}:
\begin{itemize}
\item 
\emph{An argument for (the claim) $\sent \in \LL$ 
supported by $\asmset \subseteq \A$ and $\ruleset \subseteq \RR$
}
(denoted $\argur{\asmset}{\sent}{\ruleset}$
) is a finite tree with nodes labelled by
sentences in $\LL$ or by $true$ 
(a sentence not already in $\LL$), the root labelled by $\sent$, leaves either $true$ or
assumptions in $\asmset$, and non-leaves $\sent'$ with, as children,
the elements of the body of some rule in \ruleset{} 
with head $\sent'$ (and all rules in $\ruleset$ are used in the tree).
\item 
$\argur{\asmset_1}{\sent_1}{\ruleset_1}$ 
{\em attacks} 
$\argur{\asmset_2}{\sent_2}{\ruleset_2}$  
iff
$\sent_1=\overline{\asm}$ for some
$\asm \in \asmset_2$.
\end{itemize}
%

Given a flat ABA framework \abaf, 
let $\Args$ be the set of all arguments and $\Att=\{(\alpha,\beta) \in \Args \times \Args \mid \alpha$ attacks $\beta\}$
.
Then, the notion of ``acceptable'' extensions we  will focus on 
is as follows: $\Delta\subseteq \Args$ is a \emph{stable extension} iff (i) $\nexists \alpha,\beta \!\in \!\Delta$ such that $(\alpha,\beta) \!\in \!\Att$ (i.e. $\Delta$ is \emph{conflict-free}) and (ii) $\forall \beta \!\in \!\Args\setminus \Delta, \exists \alpha \!\in \!\Delta$ such that $(\alpha,\beta) \!\in \!\Att$ (i.e. $\Delta$ ``attacks'' all arguments it does not contain, thus pre-emptively ``defending'' itself against 
attacks).
We say that an ABA framework is \textit{satisfiable} if it admits at least one stable extension, and \textit{unsatisfiable} otherwise.

Without loss of generality, 
we will leave the language component of all ABA frameworks implicit,
and use, e.g., $\sABA $  to stand for $\abaf $ where $\LL$ is the set of all sentences in $\RR$, $\A$ and in the range of~$\contrary$.
We will also write $\sABA \vdashacc_\Delta s$  to indicate 
that $\Delta$ is a stable extension of $\sABA$ and $s\in \LL$ is a claim of an argument in $\Delta$; we also say that $s$ is a \emph{brave consequence} 
 of $\sABA$.

\begin{example}
In the ABA framework $F\!=\!\sABA$ 
from Example~\ref{ex:NixonBK},
we can construct, amongst others, the following arguments:

$\argur{\emptyset}{quaker(a)}{\{\rho_1\}}$
\hfill
$\argur{\{votes\_dem(e)\}}{\textit{democrat}(e)}{\{\rho_{8},\rho_{11}\}}$

$\argur{\{votes\_rep(e)\}}{\textit{republican}(e)}{\{\rho_9,\rho_{11}\}}$

$\argur{\{normal\_quaker(a)\}}{\textit{pacifist}(a)}{\{\rho_{1},\rho_{10}\}}$

\vspace{1mm}
\noindent
with the second and third 
 arguments 
attacking each other.     
$F
$ admits 
two 
stable extensions $\Delta_1$ and 
$\Delta_2$, where
$
F\!\!\vdashacc_{\Delta_1}\!\! democrat(e)$%
and
 $
F\!\vdashacc_{\Delta_2} \!\textit{republican}(e)$.
Also, for $i\!\!=\!\!1,2$, $
F\! \vdashacc_{\Delta_i}\! \textit{pacifist}(a)$,  $
F\!\vdashacc_{\Delta_i}\! \textit{pacifist}(b)$, and $
F\!\vdashacc_{\Delta_i}\! \textit{pacifist}(e)$, 
as no argument for $\textit{abnormal}\_\textit{quaker}(X)$ can be constructed
and hence 
$\rho_{10}$ is applicable for every $X$ such that $\textit{quaker}(X)$ is accepted.


\end{example}

\section{Brave ABA Learning under Stable Extensions}
\label{sec:ABAlearning}

Here we present the instance of the ABA Learning problem that we consider in this paper. We follow the lines of \cite{MauFraILP22}, but we focus on a semantics based on brave consequences under stable extensions. Also, we consider a further parameter: the set $\mathcal T$ of predicates to be learned, which do not necessarily coincide with the predicates occurring in the sets of examples 
given in input to the learning. We will see later the role played by this set.

The \emph{background knowledge} is \emph{any} ABA framework $\sABA$.
\emph{Positive} and \emph{negative examples} are ground atoms of the form $p(c)$, for $p$ a predicate, with arity $n \geq 0$, and $c$ a tuple of $n$ constants.
Here, we impose that examples are non-assumptions (in the background knowledge $\sABA$). 
For instance, in Example~\ref{ex:NixonBK},
 $\textit{normal\_quaker}$ cannot appear in 
 examples. 
 The exclusion of assumptions from examples is derived from the flatness restriction. 
We also assume that there is a predicate $dom$ such that for every individual constant $c$, the fact $dom(c)\leftarrow$ is in~$\RR$. 



By $\textit{pred}(E)$ we denote the set of predicate symbols occurring in $E$, where $E$ is an atom, a rule, a set thereof, or an ABA framework.

\begin{definition}\label{def:ABAlearningPb}
Given a satisfiable background knowledge $\sABA$, positive examples $\mathcal{E}^+$ and
negative examples $\mathcal{E}^-$,
with $\mathcal{E}^+\cap \mathcal{E}^- =\emptyset$,
and a set $\mathcal T$ of \emph{learnable} predicates, with $\mathcal T\cap\textit{pred}(\A) =\emptyset$ and 
$\textit{pred}(\EE)\subseteq \mathcal T$,
the \emph{goal of brave ABA Learning} is to 
construct 
$\sABAp$ such that:
(i) $\RR \subseteq \RR'$, 
(ii) for each 
$H \IF B \in \RR'\setminus\RR$, $\textit{pred}(H) \cap \textit{pred}(
\sABA) \subseteq \mathcal T$,
(iii) $\A \subseteq \A'$, 
(iv)~$\overline{\alpha}'=\overline{\alpha}$ for all $\alpha \in \A$, 
(v)~$\sABAp$ is satisfiable and admits a stable extension $\Delta$, such that:
\vspace{-2mm}
\begin{enumerate}
\item for all $e \in \mathcal{E}^+$, $\sABAp\models_\Delta e$,


\item for all $e \in \mathcal{E}^-$,  $\sABAp\not\models_\Delta e$.  
\end{enumerate}

\vspace{-2mm}
\noindent
$\sABAp$ is called a \emph{solution} of the \emph{brave ABA Learning problem}  $(\sABA,$ $\EE,\mathcal T)$. We also say that $\sABAp$ \emph{bravely entails} $\EE$.
A solution is 
\emph{intensional} when $\RR' \setminus \RR$ is made out of
non-ground rule schemata.
\end{definition}

Condition (ii) requires that the head predicate of a learnt rule is either an ``old'' predicate (in 
the background knowledge or in the examples), in which case it needs to be
a learnable predicate in $\mathcal T$, or a new predicate (not already in the background knowledge, e.g. the contrary of a new assumption in $\A'\setminus \A$), in which case it does not need to be in  $\mathcal T$.
Note that we only require to specify, via $\mathcal T$, which, amongst the predicates in the background knowledge, can be subject to learning, while imposing no restrictions on 
which new predicates can be learnt, unlike existing approaches, e.g., \cite{LawRB14}.
The following example illustrates the usefulness of learning new predicates.

\begin{example}\label{ex:NixonLearningPb}
Consider the background knowledge in Example~\ref{ex:NixonBK}, and 

\smallskip

$\Ep = \{\textit{pacifist}(a),\textit{pacifist}(c),\textit{pacifist}(e) \}$,

$\En = \{\textit{pacifist}(b),\textit{pacifist}(d)\}$,

 
$\mathcal T=\{\textit{pacifist},abnormal\_quaker\}$
.

\smallskip

\noindent Solutions  of $(\sABA, \!\EE, \mathcal{T})$ include
ABA frameworks with sets of rules $\RR_1'$ and 
$\RR_2'$
whereby

\smallskip

\noindent
$\RR_1'\setminus \RR=\{abnormal\_quaker(b) \leftarrow,\quad \textit{pacifist}(c) \leftarrow \}$ 

\noindent
{$\RR_2'\setminus \RR=\{$}$abnormal\_quaker(X) \leftarrow republican(X),\ \alpha(X),$

\noindent
\phantom{$\RR_2'\setminus \RR=\{$}${c\_\alpha(X)} \leftarrow quaker(X),\ normal\_quaker(X),$

\noindent
\phantom{$\RR_2'\setminus \RR=\{$}$\textit{pacifist}(X) \leftarrow democrat(X)
\}$
\\
with $\A' = \A \cup\{\alpha(X) \mid X\in \{a,b,c,d,e\}\}$ and
$\overline{\alpha(X)}' = c\_\alpha(X)$.
%
The new assumption $\alpha(X)$ (and its contrary) is introduced by the learning algorithm we will describe in  Section~\ref{sec:LearningAlgo}, and can be interpreted as ``$X$ is a normal republican".
\end{example} 



Note that intensionality in Definition~\ref{def:ABAlearningPb} captures a notion of generality for the learnt rules, i.e., rules that do not make explicit reference to specific values in the universe.
In the former example, the second solution can be deemed to be intensional, whereas the first is not. 


The following example shows that the choice of 
$\mathcal T$ 
may affect the existence of a solution for the learning problem
. In particular, in this example, there is no solution when $\mathcal T$ is the set of predicates occurring in $\EE$, while there is a solution by taking a larger set.

\begin{example}
Consider the brave ABA Learning problem $(\sABA,\EE,\T)$, where: 
$\RR = \{p\leftarrow q, r\} $; ~ $\A = \{r\}$; \quad $\overline{r} = p$; ~  
$\Ep=\{q\}$; ~ $\En=\emptyset$; ~  $\T=\{p,q\}$.
A solution for this problem is: 
$\RR' = \{p\leftarrow q, r, ~ p\leftarrow, ~ q \leftarrow\} $; ~ $\A' = \{r\}$; ~ $\overline{r}' = p$.
However, no solution exists if we take $\T=\{q\}$, that is, if $\T$ is the set of predicates occurring in $\EE$.
\end{example}

The problem pointed out in this example may not arise if we adopt other semantics such as \textit{grounded 
} or \textit{semi-stable} extensions \cite{ABAhandbook}, instead of stable extensions. An alternative way to address this problem would be to assume that each rule in the background knowledge can be made defeasible, by adding an assumption in the body, whose contrary can be learnt.
We leave 
these lines of work for future research.




\section{Brave ABA Learning via Transformation Rules}
\label{subsec:transformation}



To learn ABA frameworks from  examples, we follow the approach based on \emph{transformation rules} 
from \cite{MauFraILP22}, but only consider a subset of those rules: \textit{Rote Learning}, \textit{Folding}, \textit{Assumption Introduction}, and (a special case of) \textit{Subsumption},
thus 
ignoring \textit{Equality Removal}~\footnote{The effect of equality removal can be obtained with the Folding rule R2 presented in this paper by replacing an equality $X=c$ with $dom(X)$, an atom that holds for all constants in the universe.}.
%
%
%
 Folding and Subsumption are borrowed from \emph{logic program transformation}~\cite{PeP94
 }, while Rote Learning and Assumption Introduction are specific for ABA.
Given an ABA framework $\sABA$, a transformation rule constructs a new ABA framework $\sABAp$ (
below, we will mention explicitly only the modified components
).



We  assume 
rules in $\mathcal{R}$ 
are written in \emph{normalised} form as follows:

\smallskip

$p_0({X}_0) \leftarrow eq_1, \ldots, eq_k, p_1({X}_1), \ldots, p_n({X}_n)$

\smallskip
\noindent
where $p_i({X}_i)$, for $0\leq i \leq n$,  is an atom (whose ground instances are) in $\mathcal L$ and $eq_i$, for $1\leq i \leq k$, is an equality $t_1^i=t_2^i$, with $t_j^i$ a term whose variables occur in the tuples ${X}_0,{X}_1, \ldots, {X}_n$.
In particular, we represent a ground fact $p(t) \leftarrow $ as $p(X) \leftarrow  {X}={t}$.
The body of a normalised rule can be freely rewritten by using the standard axioms of equality, e.g., $Y_1=a, Y_2=a$ can be rewritten as $Y_1=Y_2, Y_2=a$.
For constructing arguments, we assume that, for any ABA framework, the language $\LL$ contains all equalities between elements of the underlying universe and $\mathcal R$ includes all rules $a=a\leftarrow$, where $a$ is an element of the universe.
We also assume that, for all rules $H\IF B \in \RR$, $vars(H) \subseteq vars(B)$.
When presenting the transformation rules, we use the following notations: (1) 
$H,K$  denote heads of rules, (2) 
$Eqs$ (possibly with subscripts) denotes sets of equalities, 
(3) 
$B$ (possibly with subscripts) denotes sets of atoms. 

\smallskip
\noindent
\emph{R1. Rote Learning.} Given 
atom $p({t})$,
add ~${\rho}\!:  p({X})\leftarrow{X}\!=\!{t}$~ to $\RR$.
Thus, $\RR'\!=\!\RR\cup \{\rho\}$.

\smallskip

We will use R1 either to add facts from positive examples or facts for contraries of assumptions, as shown by the following example.

\begin{example}
\label{ex:R1} 
Let us consider the learning problem presented in Example~\ref{ex:NixonLearningPb}. By Rote Learning we add to $\RR$ the following two rules:

$\rho_{12}$: $abnormal\_quaker(X) \leftarrow X=b$

$\rho_{13}$: $\textit{pacifist}(X) \leftarrow X=c$

\noindent
The resulting ABA framework with rules $\RR \cup \{\rho_{12},\rho_{13}\}$ is a (non-intensional) solution. 
\end{example}


We will show in Section~\ref{sec:LearningAlgo} how the Rote Learning rule can be applied in an automatic way, by using ASP, so to add facts to $\RR$ that allow the derivation of a, possibly non-intensional, solution of a given ABA Learning problem.

\smallskip

\noindent
\emph{R2. Folding.}
Given distinct rules

$\rho_1$: $H \leftarrow Eqs_1, B_1, B_2$
\quad and \quad
$\rho_2$: $K \leftarrow Eqs_1, Eqs_2, B_1$

\noindent
with $\textit{vars}(\textit{Eqs}_2) \cap \textit{vars}(\rho_1)\!=\!\emptyset$, replace $\rho_1$ by 
$\rho_3$: $H \!\leftarrow\! Eqs_2, K, B_2.$
Thus, by \emph{folding $\rho_1$ using $\rho_2$}, we get $\RR'=(\RR\setminus \{\rho_1\})\cup\{\rho_3\}$.

\begin{example}
\label{ex:R2}
For instance, by folding rules $\rho_{12}$ and $\rho_{13}$ of Example~\ref{ex:R1}
using rules $\rho_{6}$ and $\rho_{4}$ (after normalisation) of Example~\ref{ex:NixonBK}, we get:

$\rho_{14}$: $abnormal\_quaker(X) \leftarrow republican(X)$

$\rho_{15}$: $\textit{pacifist}(X) \leftarrow democrat(X)$

\noindent
The resulting ABA framework whose set of rules is $\RR \cup \{\rho_{14},\rho_{15}\}$ is no longer a solution. Indeed, $\textit{pacifist}(a)$ is not a brave consequence of $\langle \RR \cup \{\rho_{14},\rho_{15}\},\A,\contrary\rangle$.
\end{example}

\smallskip

Folding can be seen as a form of {\em inverse resolution}~\cite{Muggleton1995}, 
used for generalising a rule by replacing 
some atoms in its body with their consequence using a rule in $\RR$.
In terms of logic program transformation~\cite{PeP94}, we can see that 
if we \textit{unfold} $\rho_3$ wrt $K$ using $\rho_2$ we get a rule more general than $\rho_1$. From an argumentation point of view, the following proposition shows that folding preserves arguments.

\begin{proposition}\label{prop:folding}
    Suppose that $\RR'=(\RR\setminus \{\rho_1\})\cup\{\rho_3\}$ is obtained by folding $\rho_1$ using $\rho_2$ (where $\rho_1, \rho_2, \rho_3$ are as in R2). Then, for any argument $\argur{A}{s}{R}$ with $ R \subseteq \RR$, there exists an argument $\argur{A}{s}{R'}$ with $ R' \subseteq \RR'$.
\end{proposition}

However, folding may also introduce new arguments and new attacks, and hence we have no guarantees on the preservation of extensions, as shown by the following example.

\begin{example}
Consider the ABA framework $\sABA$, where:

{$\RR = \{$}$\rho_1:~p(X)\leftarrow q(X), r(X), \quad \rho_2:~s(X) \IF X=a,$

\phantom{$\RR = \{$}$\rho_3:~s(X) \IF q(X)\} $; \quad 

$\A = \{r(X)\}$; \quad $\overline{r(X)} = p(X)$

\noindent
By folding, we get
$\RR' \!=\! (\RR \setminus \{\rho_1\}) \cup \{\rho_4:~p(X)\!\leftarrow\! s(X), r(X)\}$.

\noindent
$\RR'$ has an extra argument $\argur{\{r(a)\}}{p(a)}{\{\rho_4,\rho_2\}}$, which attacks itself, because $p(a)$ is the contrary of the assumption $r(a)$, and hence the new ABA framework does not admit any stable extension.
\end{example}

The following Assumption Introduction transformation rule adds 
an assumption to the body of a rule so to make it defeasible. By learning rules for the contrary of the assumption we may be able to avoid the acceptance of unwanted arguments.
This is particularly important during ABA Learning, e.g.,  when we want to avoid the acceptance of a negative example.

\smallskip

\noindent \emph{R3. Assumption Introduction.} 
Replace 
$\rho_1: H \leftarrow Eqs, B$
%
%
in $\RR$ 
by
$\rho_2: H \leftarrow Eqs, B, \alpha({X})$,
%
where $X$ is a tuple of variables taken from $\textit{vars}(\rho_1)$ and $\alpha({X})$ is a (possibly new)
assumption with contrary $c\_\alpha({X})$. 
Thus, $\RR'=(\RR\setminus \{\rho_1\})\cup\{\rho_2\}$, $\mathcal{A}'= \mathcal{A} \cup \{\alpha({X})\}$, $\overline{\alpha({X})}'=c\_\alpha({X})$,
and $\overline{\beta}'=\overline{\beta}$ for all $\beta \in \A$.

\begin{example}
\label{ex:R3}
By Assumption Introduction, from rule 
$\rho_{14}$ in Example~\ref{ex:R2}, we get

$\rho_{16}$: $abnormal\_quaker(X) \leftarrow republican(X), \alpha(X)$

\noindent
where, for $X\in \{a,b,c,d,e\},$ $\alpha(X)$ is an assumption with contrary $c\_\alpha(X)$.
Now, by Rote Learning we can add the fact:

$\rho_{17}$: $c\_\alpha(X) \leftarrow X=a$.

\noindent
The current ABA framework, is a (non-intensional) solution for the learning problem of Example~\ref{ex:NixonLearningPb}.
\end{example}

The ability of Assumption Introduction, together with Rote Learning, to recover a solution after Folding, as shown in Example~\ref{ex:R3}, is proved under very general conditions in the following proposition.

\begin{proposition}\label{prop:asm}
Suppose that $\langle \RR_1,\A_1,\contrary^{_1}\rangle$ is a {solution} of the brave ABA Learning problem  $(\sABAo,$ $\EE,\mathcal T)$.
Let $\RR_2=(\RR_1\setminus \{\rho_1\})\cup\{\rho_3\}$ be obtained by folding $\rho_1$ using $\rho_2$, where $\rho_1, \rho_2,$ and $\rho_3$ are as in R2.
Let $\RR_3=(\RR_2\setminus \{\rho_3\})\cup\{\rho_4\}$ be obtained by applying R3, where: $\rho_4 = H \leftarrow Eqs_2, K, B_2, \alpha({X})$, $\alpha$ is a new predicate symbol, and $vars(\{Eqs_1,B_1\})\subseteq X=vars(\rho_4)$.
Then there exists a set $S$ of atoms and a set $C_\alpha =\{c\_\alpha(X) \IF X\!=\!t \mid c\_\alpha(t)\in S\}$ of rules such that $\langle \RR_3 \cup C_\alpha,\A_1\cup\{\alpha(X)\},\contrary^{_1}\cup \{\alpha(X) \mapsto c\_\alpha(X)\}\rangle$ is a solution of $(\sABAo,$ $\EE,\mathcal T)$.
\end{proposition}

The 
transformation rule below is a variant of 
Subsumption
in
\cite{MauFraILP22}.

\smallskip
\noindent
\emph{R4. Fact Subsumption.}
Let $\EE$ be a pair of sets of positive and negative examples. Suppose that  $\RR$ contains the rule

$\rho: p(X) \leftarrow X=t$

\noindent
such that 
$\langle \RR \setminus \{\rho\},\mathcal A, \contrary \rangle $ bravely entails $\EE$.
Then, by \textit{fact subsumption relative to} $\EE$, we get $\RR'=\RR\setminus \{\rho\}$. 



\begin{example}
Let $\Ep=\{p(a)\}$, $\En=\{p(b)\}$ and consider an ABA framework with rules

$\RR =\{$$p(X) \leftarrow q(X), r(X), \ \ \ s(X)  \leftarrow q(X), t(X),$


\phantom{$\RR =\{$}$p(X) \leftarrow X\!=\!a, \ \ \ q(X)\leftarrow X\!=\!a, \ \ \ q(X)\leftarrow X\!=\!b\}$

\vspace{1mm}
\noindent
where $r(X) $,$t(X)$ are assumptions with 
$\overline{r(X)} = s(X) $ and $\overline{t(X)}=p(X)$.
Then, $\langle \RR \setminus \{p(X) \!\leftarrow\! X=a\},\mathcal A, \contrary \rangle $ bravely entails $\langle\{p(a)\},\{p(b)\}\rangle$, and hence by Fact Subsumption, the rule $p(X) \leftarrow X=a$
can be removed from $\RR$.
\end{example}

In the field of logic program transformation, 
the goal is to derive a new program that
is \emph{equivalent},
\wrt\ a semantics of choice, to the initial program.
Various results guarantee that, 
under suitable conditions, transformation rules defined in the literature, such
as Unfolding and Folding, indeed enforce equivalence (e.g., \wrt\ the least Herbrand model of definite  programs~\cite{PeP94} or the stable model semantics of normal logic programs~\cite{ArD95}).
These results have also been generalised by using argumentative notions~\cite{ToK96}.

In the context of  ABA Learning, however, program
equivalence is not a desirable objective, as we look for
sets of rules that entail, in the sense of Definition~\ref{def:ABAlearningPb}, given sets of positive and negative examples.
We will show in the next section how suitable sequences of applications of the transformation rules can be guided towards the goal of computing a solution of a given brave ABA Learning problem.

\section{A Brave ABA Learning Algorithm}\label{sec:LearningAlgo}


The application of the transformation rules is guided by the 
$\Babalearn$ algorithm (see Algorithm~\ref{alg:strategy}),
a variant of the one in~\cite{DPT23a}, which refers to a cautious 
stable extensions semantics.
The goal of $\Babalearn$ is to derive an intensional solution 
for the given brave ABA Learning problem, and to achieve that goal 
some tasks are implemented via an ASP solver.

The $\Babalearn$ algorithm is the composition of two 
procedures~\textit{RoLe} and~\textit{Gen}: 

\noindent
(1)~\textit{RoLe} repeatedly applies Rote Learning with the objective of
adding a minimal set of facts to the background knowledge $\sABAo$
so that the new ABA framework $\sABA$ is a (non-intensional) solution 
of the brave ABA Learning problem $(\sABAo, \EE, \mathcal T)$ given in input.

\noindent
(2)~\textit{Gen} has the objective of transforming $\sABA$ into an intensional solution. This is done by transforming each learnt non-intensional rule as follows. First, \textit{Gen} repeatedly applies Folding, so as to get a new intensional rule. It may happen, however,  that the ABA framework with the new rule is no longer a solution of the given brave ABA Learning problem, because, as mentioned in the previous section, new arguments and attacks may be added. In this case, \textit{Gen} applies Assumption Introduction, followed by Rote Learning (that is, finds suitable exceptions to the learnt rules), and derives a new ABA framework that is a solution (as guaranteed by Proposition~\ref{prop:asm}). 
Then, redundant facts are removed by Fact Subsumption.
\textit{Gen} is iterated until all learnt rules are intensional, or a failure to compute a solution is reported.

Both \textit{RoLe} and \textit{Gen} exploit the existence of a mapping between ABA 
frameworks under the stable extension semantics and ASP programs~\cite{ABA}, and make use of the 
following encoding into ASP rules of a given ABA Learning problem (we use the teletype 
font for ASP rules).

\begin{definition}\label{def:ASPenc}
Let $\mathtt {dom(t)}$ hold for all tuples $\mathtt{t}$ of constants of $\LL$.  
We denote by $ASP(\sABA, \EE, \mathcal T)$ the set of ASP rules constructed as described at the following points (a)--(e).

\begin{enumerate}[label=(\alph*),leftmargin=15pt,topsep=0pt]
\item Each rule in $\mathcal R$ is rewritten in the ASP syntax (see Section~\ref{sec:background}). 

\item Each $\alpha\!\in\!\A$ occurring in $\RR$ is encoded as the following ASP rule, 
where $\mathtt c\_\alpha$ is an ASP atom encoding $\overline{\alpha}$, 
and $vars(\mathtt{\alpha}) = \mathtt{X}$:

\quad ${\tt \alpha\pif~dom(X),~not~c\_\alpha.}$

\item Each $e\in\Ep$ is encoded as the ASP rule 
${\tt\pif ~not~e.}$

\item Each $e\in \En$ is encoded as the ASP rule 
${\tt\pif ~e.}$

\item Each atom $p(X)$ with $p \in \mathcal T$ is  encoded through 
the following ASP rules, where $\mathtt {p'}$ is a new predicate name:

\quad ${\tt p(X)\pif ~p'(X).}$ \quad\quad $\mathtt{\{p'(X)\}\pif~dom(X).}$

with the ASP directive ${\tt{\#minimize \{ 1,X: p'(X) \}}}$.

\end{enumerate}
\end{definition}

The rule $\mathtt{\{p'(X)\}\pif~dom(X)}$ at point 
(e) is a \emph{choice rule}~\cite{Ge&12}, 
which has an answer set for each subset of 
$\{{\tt p(t)\mid dom(t)}$ holds$\}$\footnote{
In the implementation of Algorithm~\ref{alg:strategy} we make use of the following optimisation, which reduces the domain size of each variable in $\mathtt X$, and hence the size of the grounding of the choice rule.
If $p\in\mathcal{T}$ is the predicate of a contrary of an  assumption $\alpha(X)$ and $B$ is the body in which $\alpha(X)$ occurs, $\mathtt{dom(X)}$ is replaced by the conjunction of the non-assumption atoms $b$ in $B$ such that  $\textit{vars}(\mathtt{X})\cap\textit{vars}(b)\not=\emptyset$.
If $p$ occurs in $\Ep$, the choice rule is simplified to 
``$\mathtt{\{e_1 ;\dots; e_n\}}.$'' where
${\{e_1,\dots,e_n\}}=\{ p'(t)\mid p(t) \in \Ep\}$.}.
We use predicate $\mathtt{p'}$ to distinguish new facts from the
atoms $\mathtt{p(X)}$ which are already consequences of the ASP rules (a)--(d). 
The ${\tt{\#minimize \{ 1,X : p'(X) \}}}$ directive does not affect satisfiability, but enforces the computation of answer sets 
with \textit{minimal} subsets of ${\tt p'}$ atoms, and hence the addition of a minimal set of new facts by \textit{RoLe}. 




The following properties of $ASP(\sABA, \EE, \mathcal T)$ 
will be used for showing the soundness of the 
$\Babalearn$ algorithm.

\begin{theorem}
    \label{thm:entails}
$\sABA$ bravely entails $\EE$ if and only if the set of rules $ASP(\sABA,\EE,\emptyset)$ is satisfiable.    
\end{theorem}

Thus, in particular, a claim $s\in\LL$ is a brave consequence of $\sABA$ under stable extensions if and only if the set of rules $ASP(\sABA, \langle \{s\}, \emptyset \rangle, \emptyset)$ is satisfiable.

\begin{theorem}\label{thm:sol}
(1) There exists a solution of the brave learning problem $(\sABA,$ $\EE,\mathcal T)$ if and only if the set of rules $ASP(\sABA,\EE,\mathcal T)$ is satisfiable.
(2) Suppose that $S$ is an answer set of $ASP(\sABA,\EE,\mathcal T)$, then $\langle \RR',\A,\contrary\rangle$ is a solution if ~$\RR'=\RR\cup\{p(X) \IF X=t \mid p\in \mathcal T$ and $\mathtt {p'(t)} \in S\}$.
\end{theorem}


\input{algorithm}

Let us comment the $\Babalearn$ algorithm in some detail.
At line \ref{alg:r1} of the \textit{RoLe} procedure, the algorithm sets to $P$ the ASP encoding $ASP(\sABA,\EE,\mathcal T)$ of the input learning problem.
The failure at line \ref{alg:r2} is due to Theorem \ref{thm:sol}:
if $P$ is unsatisfiable, then the learning problem has no solutions. 
Otherwise, due to the ${\tt{\#minimize \{1,X:p'(X)\}}}$
directive, the answer set $S$ computed by function $\textit{getAS}(P)$ 
at line \ref{alg:r3} will contain a \textit{minimal} set of 
new atoms that can be learnt (see lines \ref{alg:r4}--\ref{alg:r6})
so to obtain a (non-intensional) solution. 
This directive will also minimize the set of alternative answer sets
that are computed at line \ref{alg:r3} in case of backtracking.

At line \ref{alg:g1} of the \textit{Gen} procedure, 
the algorithm considers any non-intensional rule $\rho\!\in\!\RRl$
of the form $p(X)\!\leftarrow\!X\!=\!t$.
At line~\ref{alg:g14} the algorithm applies Fact Subsumption 
to $\rho$, that is, it checks whether or not it can be deleted 
by preserving the brave entailment of $\EE$.
If this is not the case, at line \ref{alg:g2}, it applies the
function $\textit{applyFolding}$, which is defined in a way 
that (see Definition~\ref{def:applyFolding}),
given a non-intensional rule $\rho$, it returns an intensional
rule $\rho_f$ obtained by applying once or more times (possibly, 
in a nondeterministic fashion) the Folding transformation using
rules in $\RRl\setminus\{\rho\}$.

The applications of the Folding transformation may derive an ABA framework that is no longer a solution of the given learning problem, because the examples may no longer be entailed. Indeed, at line \ref{alg:g3}, the algorithm checks if the current ABA framework 
$\langle\RR\cup\RRl\!\cup\!\{\rho_f\},\A,\!\contrary\rangle$
bravely entails $\EE$, that is, by Theorem~\ref{thm:entails}, if 
$ASP(\langle\RR\cup\RRl\!\cup\!\{\rho_f\},\A,\!\contrary\rangle,\!\langle\Ep\!,\En\rangle,\emptyset)$ 
is satisfiable. If it is not, by the following lines 
\ref{alg:g5}--\ref{alg:g13}, the ABA framework is transformed into a  
(possibly non-intensional) solution.

The first step to get again an ABA framework 
that is a solution of the input learning problem 
is to apply Assumption Introduction. 
This is done at line \ref{alg:g5}, using the function defined at lines
\ref{alg:asm}--\ref{alg:a10}, 
where the algorithm may either 
(lines \ref{alg:a2}--\ref{alg:a5}) take an $\alpha(X)$ in the set $\A$ 
or 
(lines \ref{alg:a6}--\ref{alg:a9}) 
introduce a new assumption. 
This choice is a key point for enforcing the termination of the algorithm,  as using an assumption in $\A$ may avoid the introduction of an unbounded number of new predicates.  

In the case where the algorithm uses an assumption $\alpha(X)$  already belonging to $\A$ (see line \ref{alg:a2} and Definition~\ref{def:univocal}) and it does not obtain a solution, then it gets a failure 
(see line \ref{alg:a5}) and backtracks to the most recent choice point. 
This point can be line \ref{alg:g5}, if \textit{applyAsmIntro} is nondeterministic (that is, the choice of the assumption $\alpha(X)\in\mathcal{A}$ at line \ref{alg:a2} is nondeterministic), or line \ref{alg:g2}, if \textit{applyFolding} is nondeterministic. 
In the case where no alternative choice is possible at lines  \ref{alg:g2} and \ref{alg:g5}, the algorithm halts with failure. 

If at line \ref{alg:a6} the algorithm introduces a new assumption, then at line \ref{alg:g6}, it updates the set of assumptions and their contraries. 
In the case where a new $\alpha$ is introduced, Proposition~\ref{prop:asm} guarantees that the answer set $S$ to be computed at line \ref{alg:a9} exists and, 
by adding the facts for its contrary $c\_\alpha$ via Rote Learning
(see lines \ref{alg:g12}--\ref{alg:g13}), we always get a 
non-intensional solution for the given brave ABA Learning
problem.


We do not provide a concrete definition of the function \textit{applyFolding}, but we require that it satisfies the conditions specified by the following definition, where $bd(\rho)$ denotes the body of rule $\rho$.

\begin{definition}
\label{def:applyFolding}
Let $\textit{applyFolding}(\rho,\RR)$ be defined as in Algorithm~\ref{alg:strategy} with two subsidiary functions:
(i) $\textit{foldable}(\rho,\RR)$, 
a boolean-valued function such that $\textit{foldable}(\rho,\RR)$ implies that $R2$ can be applied to $\rho$ using a rule in $\RR$,
and 
(ii) $\textit{fold}(\rho,\RR)$, 
such that $\textit{fold}(\rho,\RR)$ is obtained by applying $R2$ to $\rho$ (possibly, in a nondeterministic way).
For a sequence
$\rho_0,\rho_1,\dots$ of rules such that $\rho_{i+1}=\textit{fold}(\rho_i,\RR)$, for $i\geq 0$, we define a sequence of sets of atom:

$\mathbb B_0 = bd(\rho_0)$

$\mathbb B_{i+1} = \mathbb B_i \cup \{Eqs_2, K\}$

\noindent
where $B_1, Eqs_1, Eqs_2, K$ are as in R2 and $\{B_1, Eqs_1\} \subseteq \mathbb B_i$.
We say that \textit{applyFolding} is \emph{bounded} if the following conditions hold:

\smallskip
\begin{itemize}[leftmargin=18pt, noitemsep, topsep=0pt]
\item[F1.] if $\rho$ is non-intensional,
then $\textit{foldable}(\rho,\RR)$ is \textit{true};


\item[F2.] for all $i\geq 0$, if $\textit{foldable}(\rho_i,\RR)$, then
(i) $K\not\in \mathbb B_i$, and
(ii) for any variables $X,Y,$ and constant $a$, if $X\!=\!a \in Eqs_2$, then $Y\!=\!a \not\in \mathbb B_{i}$.
\end{itemize}


\end{definition}


A simple definition of $\textit{foldable}(\rho,\RR)$ is, for instance, a function which holds true 
for any $\rho$ that has an occurrence of an equality $X=a$ in its body, where $a$ is a constant,
and for every constant $c$ in $\LL,$ there is a rule $dom(X) \IF X\!=\!c$ in $\RR$.
This function will enforce the termination of \textit{applyFolding} when no constants occur in~$\rho$.

Conditions  $F1,F2$ ensure that each application of $\textit{applyFolding}$ terminates, and its output is an intensional rule. 

\begin{proposition}\label{prop:foldterm}
Suppose that $\textit{applyFolding}(\rho,\RR)$ satisfies the conditions of Definition~\ref{def:applyFolding}.
Then, $\textit{applyFolding}(\rho,\RR)$ terminates and returns an intensional rule.
\end{proposition}

The language of the ABA framework can be extended by introducing new assumptions and their contraries, and  
an unbounded introduction of new predicates is a possible source of nontermination of Algorithm~\ref{alg:strategy}. 
The following definition of an \textit{assumption relative to} a rule body will be used to enforce the introduction of a bounded set of assumptions
and contraries.

\begin{definition}\label{def:univocal}
Suppose that $\rho= H \leftarrow B, \alpha(X)$ is a rule in $\RR$
where $\alpha(X)\in \A$.
Then we say that $\alpha(X)$ is an \textit{assumption relative to} $B$. 
\end{definition}


We say that Algorithm $\Babalearn$ \textit{terminates with success} for a given brave ABA Learning problem if it halts and returns a solution. If it halts and does not return a solution, then it \textit{terminates with failure}.
Putting together the results proved above, we get the soundness of Algorithm~\ref{alg:strategy}.

\begin{theorem}[Soundness]\label{thm:soundness}
    If Algorithm $\Babalearn$ terminates with success for an input brave ABA Learning problem, then its output is an intensional solution.
\end{theorem}





Now, by Proposition~\ref{prop:foldterm} and the fact that, at line \ref{alg:a2} of Algorithm~\ref{alg:strategy}, \textit{applyAsmIntro} uses an assumption in the current set $\A$,
whenever in $\A$ there exists an assumption relative to the body of the rule under consideration, we get the termination of $\Babalearn$.

\begin{theorem}[Termination] Suppose that \textit{applyFolding} is bounded (see Definition~\ref{def:applyFolding}). Then Algorithm $\Babalearn$  terminates (either with success or with failure).
\end{theorem}


\begin{example}
    The applications of the transformation rules shown in previous examples 
    for the brave ABA learning problem of Example~\ref{ex:NixonLearningPb} can be seen as applications of Algorithm~\ref{alg:strategy}. 
    Indeed, Example~\ref{ex:R1} 
    shows an application of
     \textit{RoLe} (lines \ref{alg:role}--\ref{alg:endrole}), Example~\ref{ex:R2} shows two applications (in 
    different iterations) of \textit{applyFolding} (line~\ref{alg:g2}) to rules obtained by \textit{RoLe}, and Example~\ref{ex:R3} 
    shows an application of applyAsmIntro (line~\ref{alg:asm}) and the subsequent Rote Learning of the facts computed by ASP for the contrary  $c\_\alpha(X)$ (lines \ref{alg:g12}--\ref{alg:g13}).
    The learning algorithm continues from rule $\rho_{17}$ of Example~\ref{ex:R3} by performing a new iteration of Procedure $\textit{Gen}$. Function \textit{applyFolding} (line~\ref{alg:g2}) gets:

    $\rho_{18}$: $c\_\alpha(X) \IF quaker(X)$

    \noindent
    The satisfiability test (line \ref{alg:g3}) fails, as the new ABA framework is not a solution. Now, the \textit{applyAsmIntro} function proceeds by looking for an assumption relative to  $quaker(X)$ in the current set of assumptions (line \ref{alg:a2}). This assumption is $normal\_quaker(X)$ and, indeed, by replacing $\rho_{18}$ with

    $\rho_{19}$: $c\_\alpha(X) \IF quaker(X), normal\_quaker(X)$

    \noindent
    we get an ABA framework that is an intensional solution of the given learning problem (this solution coincides with the ABA framework including rules $\RR_2'$ in Example~\ref{ex:NixonLearningPb}).
    The learnt ABA framework has (among others) a stable extension including the 
    arguments:

\hspace{1.8mm}$\argu{\emptyset}{quaker(a)},$~~~~ $\argu{\{normal\_quaker(a)\}}{\textit{pacifist}(a)},$
   
   

\hspace{1.8mm}$\argu{\{\alpha(b)\}}{abnormal\_quaker(b)},$


\hspace{1.8mm}$\argu{\emptyset}{\textit{democrat}(c)},~~~~ \argu{\emptyset}{\textit{pacifist}(c)},$~~~~  $\argu{\emptyset}{republican(d)},$

\hspace{1.8mm}$\argu{\emptyset}{quaker(e)},$~~~~ $\argu{\{normal\_quaker(e)\}}{\textit{pacifist}(e)}$.

\noindent
Note that there are other stable extensions of the resulting ABA framework where, however, either $a$ is not \textit{pacifist} or $b$ is \textit{pacifist}, and thus cautious reasoning would not~work.
\end{example}


Algorithm $\Babalearn$ may terminate with failure 
in the case where \textit{applyAsmIntro} uses an
assumption $\alpha(X)$ already in the current set $\A$ (line~\ref{alg:a2}), the resulting rule $\rho$:~$H\!\leftarrow\!B, \alpha(X)$ 
does not produce a solution (line~\ref{alg:a4}), and no alternative application of Folding is available when backtracking to line~\ref{alg:g2}.
Algorithm $\Babalearn$ may halt with failure even in cases where a solution exists, but computing it would require, for instance, introducing 
other assumptions. In this sense $\Babalearn$ is not complete.
However, the learning algorithm can be slightly modified so that it always terminates with success whenever the learning problem has a solution, possibly returning a non-intensional solution. This modification is realised by allowing $\textit{applyFolding}(\rho,\RR)$ to return $\rho$ after having tried unsuccessfully 
all possible applications of the Folding transformation. Thus, the resulting rule will be non-intensional.
The modified algorithm is called $\EBabalearn$ ($E$ stands for \textit{Enumerating}).

\begin{theorem}[Soundness and Completeness of $\EBabalearn$]\label{thm:completeness}
For all brave ABAlearn problems, 
$\EBabalearn$ terminates and returns a, possibly non-intensional, solution, if a solution exists.
\end{theorem}

\section{Implementation and Experiments}

We have realised a proof-of-concept implementation~\cite{ecai2024codedata} 
of our $\Babalearn$ strategy using 
the SWI-Prolog system\footnote{SWI-Prolog v9.0.4, \url{https://www.swi-prolog.org/}} and 
the Clingo ASP solver\footnote{Clingo v5.6.2, \url{https://potassco.org/clingo/}}. 
We have used Prolog as a fully fledged programming language to handle symbolically 
the rules and to implement the nondeterministic search for a solution to the learning 
problem, while we have used ASP as a specialised solver for computing answer sets 
corresponding to stable extensions.
In particular, our tool consists of two Prolog modules implementing \textit{RoLe} 
and \textit{Gen} and two further modules implementing 
(i) the $ASP$ encoding of Definition~\ref{def:ASPenc}, and 
(ii) the API to invoke Clingo from SWI-Prolog 
and collect the answer sets to be used by \textit{RoLe} and~\textit{Gen}. 

Table~\ref{tab:exp} reports the results of the experimental
evaluation we have conducted on a benchmark set consisting 
of seven classic learning problems taken from the literature 
(\textit{flies}~\cite{DimopoulosK95},
\textit{innocent}~\cite{ABA},
\textit{nixon\_diamond}~\cite{nixon}, and variants thereof), 
and three larger problems (i.e., tabular datasets) 
from~\cite{foldrm}, to show that our approach works for non-trivial, 
non-ad-hoc examples. 
The discussion on scalability is out of the scope of the present paper.

In the table, we compare $\Babalearn$ with ILASP, a state-of-the-art 
learner for ASP programs\footnote{ILASP v4.4.0, using option 
\texttt{--version=2}, \url{https://doc.ilasp.com/}}.
When running ILASP we have opted for adopting the most direct 
representations of the learning problems, in terms of mode declarations.
\footnote{The $\Babalearn$ and ILASP specifications are archived 
in~\cite{ecai2024codedata}.}
In the ILASP column, \textit{unsat} indicates that the system
halted within the timeout, but was unable to learn an ASP program. 
These \textit{unsat} results are due to the fact that the predicates
and the mode declarations specified in the background knowledge are 
not sufficient to express a solution for the learning problem.
In this class of problems, the use of Assumption Introduction proposed 
in this paper may demonstrate its advantages.
Indeed, for instance, we have also tried ILASP on the $acute$ dataset 
extending the background knowledge with additional information matching 
the use of assumptions and their contraries automatically introduced by 
$\Babalearn$.
This manual addition has allowed ILASP to learn a solution in about 37 seconds. 

We refrained from comparing $\Babalearn$ with tools like 
FOLD-RM~\cite{ShakerinSG17,foldrm}, which can only learn \emph{stratified} normal logic programs.
These programs admit a single stable model and brave learning is not significant for them.

\begin{table}[h]
\caption{Experimental results. 
Experiments have run on an Apple M1 equipped with 
8 GB of RAM, setting a \textit{timeout} of 15 minutes.
Times are in seconds.
Column $\Babalearn$ reports the sum of the CPU and System 
times taken by out tool to compute a solution for the 
Learning problem, while column ILASP reports the time 
taken by ILASP system log.
Columns \textit{BK}, $\Ep$, and $\En$ report the number 
of rules in the background knowledge, and the number of 
positive and negative examples, respectively.
}
\vspace{12pt}
\label{tab:exp}
\centering
\begin{tabular}{@{\hspace{2mm}}l@{\hspace{3mm}}r@{\hspace{3mm}}r@{\hspace{3mm}}rrr} 
 \hline
 \toprule
 Learning problem & \textit{BK} & $\Ep$ & $\En$ & $\Babalearn$ & ILASP\\[0.5pt]
 \hline\\[-7pt] 
 \textit{flies} & 8 & 4 & 2 & 
 0.01
 & 0.09 \\ 
 \textit{flies\_birds\&planes} & 10 & 5 & 2 & 
 0.02
 & 0.25 \\ 
 \textit{innocent} & 15 & 2 & 2 & 
 0.01
 & 1.84 \\ 
 \textit{nixon\_diamond}    &  6 & 1 & 1 & 
 0.01
 & \textit{unsat}  \\ 
 \textit{nixon\_diamond\_2} & 15 & 3 & 2 & 
 0.01
 & \textit{unsat} \\ 
 \textit{tax\_law}    & 16 & 2 & 2 & 
 0.02
 & 0.66 \\ 
 \textit{tax\_law\_2} & 17 & 2 & 2 & 
 0.01
 & 0.92 \\[1pt] 
 \hline\\[-7pt] 
 \textit{acute}    &   96 &  21 &  19 &   
 0.04
 & \textit{unsat} \\ 
 \textit{autism}   & 5716 & 189 & 515 & 
 23.43
 & \textit{timeout} \\ 
 \textit{breast-w} & 6291 & 241 & 458 & 
 16.32
 & \textit{timeout} \\
 \bottomrule
\end{tabular}
\end{table}

\section{Conclusions}
\label{sec:concl}

We have designed an approach for learning ABA frameworks based on transformation rules~\cite{MauFraILP22}, and we have shown that, in the case of brave reasoning under the stable extension semantics, many of the reasoning tasks used by that strategy can be implemented through an ASP solver. 
We have studied a number of properties concerning both the transformation rules and an algorithm that implements our learning strategy, including its termination, soundness, and completeness, under suitable conditions.
A distinctive feature of our approach is that argumentation plays a key role not only at the representation level, as we learn
defeasible rules represented by ABA frameworks, but also at the meta-reasoning level, as our learning strategy can be
seen as a form of debate that proceeds by conjecturing general rules that cover the examples and then finding exceptions to them.

Even if the current implementation is not optimised, it allows solving some non trivial learning problems.
The most critical issue is that the application of 
Folding
, needed for generalisation, is non-deterministic, as there may be different choices for the rules to be used for applying 
it. 
Currently, 
we are experimenting various
mechanisms to control Folding for making it more deterministic. 
%
In addition to refining the implementation, we are also planning to perform a more thorough experimental comparison with non-monotonic ILP systems (such as FOLD-RM~\cite{ShakerinSG17,foldrm} and ILASP~\cite{LawRB14}).
%
%
Further extensions of our ABA Learning approach can be envisaged, exploiting the ability of ABA frameworks to be instantiated to different logics and semantics, and possibly address the problem of learning non-flat ABA frameworks~\cite{ABA,ABAhandbook}.
To this aim we may need to integrate ABA Learning with 
tools that go beyond ASP solvers (e.g. \cite{tuomo}).



\clearpage

\begin{ack}
We thank support from the Royal Society, UK
(IEC\textbackslash R2\textbackslash 222045 - International Exchanges 2022).
Toni was partially funded by the ERC under
the EU’s Horizon 2020 research and innovation 
programme (grant agreement No. 101020934) and 
by J.P. Morgan and the Royal Academy of Engineering,
UK, under the Research Chairs and Senior Research Fellowships scheme.
This paper has also been partially supported by the 
Italian MUR PRIN 2022 Project DOMAIN 
(2022TSYYKJ, CUP B53D23013220006, PNRR M4.C2.1.1) funded by the European Union – NextGenerationEU and by the PNRR MUR project PE0000013-FAIR.
De Angelis and Proietti are members of the INdAM-GNCS research group.
\end{ack}


\bibliography{ABA-learn}

\clearpage


\appendix

\begin{center}
{\Large \bf 
  Appendix
}
\end{center}    





\section*{Proof of Proposition~\ref{prop:folding}}
By induction on the structure of $\argur{A}{s}{R}$. 
    
\noindent
Case 1: $\rho_1$ is not the rule used at the root of $\argur{A}{s}{R}$. 
Let $\rho$: $s \IF s_1,\ldots,s_m$ be the ground rule (where $\rho$ is not an instance of $\rho_1$)
used to construct the root $s$ from the arguments $\argur{A_1}{s_1}{R_1}$, \ldots, $\argur{A_m}{s_m}{R_m}$.
By the inductive hypothesis,  for $i=1,\ldots,m$, there exists $\argur{A_i}{s_i}{R'_i}$,
with $R'_i\subseteq \RR'$. 
Thus, $\argur{A}{s}{R'_1\cup \ldots \cup R'_m}$,
with $R'_1\cup \ldots \cup R'_m \subseteq \RR'$.

\noindent
Case 2: $\rho_1$ is the rule used at the root of $\argur{A}{s}{R}$. 
Let $\rho$: $H' \IF Eqs'_1,B'_1,B'_2$ be the ground instance of rule $\rho_1$
with $H'=s$,  $Eqs'_1$ ground identities, $B'_1 = s_1,\ldots,s_k$, and $B'_2 =  s_{k+1},\ldots,s_{m}$.
Let $\argur{A_1}{s_1}{R_1}$, \ldots, $\argur{A_m}{s_m}{R_m}$ be the sub-trees of
$\argur{A}{s}{R}$ rooted at $s_1,\ldots,s_m$, respectively.
By the inductive hypothesis,  for $i=1,\ldots,m$, there exists $\argur{A_i}{s_i}{R'_i}$,
with $R'_i\subseteq \RR'$.
An argument $\argur{A}{s}{R'}$, with $R'\subseteq \RR'$ can be constructed by using
(1)~a ground instance $H' \IF Eqs'_2,K',B'_2$ of $\rho_3$ at the top,
(2)~a ground instance $K'\IF Eqs'_1, Eqs'_2, B'_1$ of $\rho_2$ (in R2 we require
$\rho_1\neq\rho_2$, and hence $\rho_2 \in \RR'$), and
(3)~for $i=1,\ldots,m$, the arguments  $\argur{A_i}{s_i}{R'_i}$ for $B'_1,B'_2$.
The condition $\textit{vars}(\textit{Eqs}_2) \cap \textit{vars}(\rho_1)=\emptyset$
allows the construction of the suitable ground instances at points (1) and (2),
because the equalities $Eqs_2$ do not add extra constraints on the variables of
$H$, $Eqs_1$ and $B_2$.

\section*{Proof of Proposition \ref{prop:asm}}
    
By the definition of a solution of a brave ABA Learning problem, $\langle \RR_1,\A_1,\contrary^{_1}\rangle$ admits a stable extension $\Delta$ such that, for all $e\in \Ep$, $\langle \RR_1,\A_1,\contrary^{_1}\rangle \models_\Delta e$ and,
for all $e\in \En$, $\langle \RR_1,\A_1,\contrary^{_1}\rangle \not\models_\Delta e$.
Let us consider the rule $\rho'_4$:  $H\IF Eqs_1, Eqs_2, K, B_1, B_2, \alpha(X)$ obtained from $\rho_1$ by adding $Eqs_2, K, \alpha(X)$ to its body, where $\textit{vars}(\{Eqs_1,B_1\})\subseteq X$, and let $R'_4=(R_1\setminus\{\rho_1\})\cup \{\rho'_4\}$.
By induction on the argument structure, for all $s,$ $\argur{A_1\!}{s}{R_1\!\!\!} \in \Delta$ iff $\argur{A_1\cup\{\alpha(X)\}\!}{\!s}{R'_4\!} \in \Delta$ (note, in particular, that no rule for the contrary $c\_\alpha(X)$ of $\alpha(X)$ occurs in $R'_4$).
Let $B_1= s_1,\ldots,s_k$, let $S=\{c\_\alpha(t) \mid Eqs_1\{X/t\}$ \textit{is a set of identities and }
$\exists i\in\{1,\ldots,k\}.\ A_1\cup\{\alpha(X)\}\not\models_{R'_4}{s_i}\}$, let $C_\alpha$ be defined from $S$ as 
in the statement of this proposition, and let $R_4=R_3\cup C_\alpha$ (that is, $R_4$ is obtained from $R'_4$ by dropping 
$Eqs_1,B_1$ from the body of $\rho'_4$ and adding to the resulting set of rules the facts for the contrary of $\alpha(X)$).
Now, we can construct a new extension $\Delta'$ such that, for all claims $s\neq c\_\alpha(X),$ $\argur{A_1}{s}{R_1} \in \Delta$ iff $\argur{A'_1}{s}{R_3} \in \Delta'$, with $A'_1\subseteq (A_1\cup\{\alpha(X)\}) \cap \overline{S}$.
Let $\mathcal F_4 = \langle \RR_4,\A_1\cup\{\alpha(X)\},\contrary^{_1}\cup \{\alpha(X) \mapsto c\_\alpha(X)\}\rangle$.
Then, $\Delta'$ is a stable extension such that, for all $e\in \Ep$, $\mathcal F_4 \models_{\Delta'} e$ and, for all $e\in \En$, $\mathcal F_4 \not\models_{\Delta'} e$, and hence $\mathcal F_4$ is a solution of $(\sABAo,$ $\EE,\mathcal T)$.

\section*{Proof of Theorem \ref{thm:entails}}

Let us consider the ASP program $P_{\mathcal F}$ constructed from the ABA framework $\mathcal F=\sABA$ by 
replacing every assumption $\alpha$ occurring in the body of a rule in $\RR$ by the negation-as-failure literal ${\tt not~\overline{\alpha}}$.
By Theorem 3.13 of ~\cite{ABA}, there is a one-to-one correspondence between the stable extensions
of $\mathcal F$ and the answer sets of $P_{\mathcal F}$. 
By rewriting each ASP rule of $P_{\mathcal F}$ of the form ${\tt p \pif~ B, not~\overline{\alpha}}$ into the pair of rules 
${\tt p \pif~ B, \alpha, ~\alpha \pif~ dom(X), not~\overline{\alpha}}$, where $vars(\mathtt{\alpha}) = \mathtt{X}$, 
we get a new ASP program $P'_{\mathcal F}$, consisting of the rules of points (a) and (b) of Definition~\ref{def:ASPenc}.
This rewriting preserves answer sets, and hence, their one-to-one correspondence with stable extensions.
For every, stable extension $\Delta$ of $\mathcal F$, let us denote by $\Delta_{\textit{as}}$ the corresponding answer set of 
$P'_{\mathcal F}$. For every claim $s$ in the language of $\LL$, $\mathcal F \models_{\Delta} s$ iff 
$s \in $ $\Delta_{\textit{as}}$. By adding the rules of points (c) and (d) of Definition~\ref{def:ASPenc}
we get a new ASP program $P''_{\mathcal F}$ that is satisfiable iff there is an answer set $\Delta''_{as}$
such that, for each $e\in\Ep$, ${\tt e} \in \Delta''_{as}$, and for each $e\in \En$, ${\tt e} \not\in \Delta''_{as}$.
Thus, $P''_{\mathcal F}$ is satisfiable iff for each $e\in\Ep$, $\mathcal F \models_{\Delta''} e$, and for each 
$e\in \En$, $\mathcal F \not\models_{\Delta''}e$. The thesis follows from  Definition~\ref{def:ABAlearningPb}, by
observing that $P''_{\mathcal F} = ASP(\sABA,\EE,\emptyset)$.

\section*{Proof of Theorem \ref{thm:sol}}

Part (1). Similarly to the proof of Theorem~\ref{thm:entails}, we make use of the one-to-one correspondence between the stable extensions of ABA frameworks and answer sets of the corresponding ASP programs, taking into account that the head predicate of a learnt rule must either occur in $\mathcal T$ or be a predicate not occurring in $\sABA$.
Part (2). If $S$ is an answer set of $P=ASP(\sABA,\EE,\mathcal T)$, then (by the rules at point (e) of Definition~\ref{def:ASPenc}) 
$P\cup \{\mathtt {p(t).} \mid  p\in \mathcal T$ and $\mathtt {p'(t)} \in S\}$ is satisfiable.
Thus, by Theorem~\ref{thm:entails}, $\langle \RR',\A,\contrary\rangle$, with $\RR'=\RR\cup\{p(X) \IF X=t \mid p\in \mathcal T$ and $\mathtt {p'(t)} \in S\}$,
 is a  solution of the given brave ABA Learning problem.

\section*{Proof of Theorem \ref{thm:soundness}}

\noindent
Suppose that a solution (either intensional or non-intensional) of the input brave ABA Learning problem $(\sABAo,\EE,\T)$ exists.
Then, by Theorem~\ref{thm:sol} there exists one that can be computed by Rote Learning using Procedure \textit{RoLe}. By Theorem~\ref{thm:sol} (2), the facts $\RR_l$ learned by \textit{RoLe} (lines \ref{alg:r4}--\ref{alg:r6}) are computed as an answer set $S$ (line \ref{alg:r3}) of the ASP rules of Definition 2 (i.e., $\mathit{ASP}(\sABAo,\EE,\T)$). (These facts will have predicates in $\mathcal T$ and may include a subset of the positive examples $\Ep$.) 
Thus, when Procedure \textit{Gen} is called, $\langle\RR\cup\RR_l,\A,\contrary\rangle$, with $\RR=\RRin$, 
$\A= \A_0$, and $\contrary = \contrary^{_0}$, is a (non-intensional) solution of $(\sABAo,\EE,\T)$.
This property is an invariant of the foreach-loop of \textit{Gen}: at each iteration of the body of the foreach-loop of \textit{Gen}, at line \ref{alg:g1}, 
the learned ABA framework $\langle\RR\cup\RR_l,\A,\contrary\rangle$ is a (possibly non-intensional) solution. 
The invariant is preserved by Fact Subsumption, as $\rho$ is deleted and not considered for Folding only if
$\textit{sat}(\mathit{ASP}(\langle(\RR\cup\RR_l)\setminus \{\rho\},\A,\!\contrary\rangle,\EE,\emptyset))$, that is, 
$\langle (\RR\cup\RR_l)\setminus \{\rho\},\A,\!\contrary\rangle$ bravely entails $\EE$.

After learning a new rule $\rho_f$ by Folding (line \ref{alg:g2}), either $\Babalearn$ gets a solution, that is, $\textit{sat}(
  \mathit{ASP}(\langle 
    \RR \cup \RR_l \cup \{\rho_f\}, \A,\!\contrary \rangle,
  \langle\Ep\!,\En\rangle,\emptyset))$ (line \ref{alg:g3}), and hence the invariant is preserved, or
  the learning algorithm proceeds by Assumption Introduction (line \ref{alg:g5}).

If \textit{applyAsmIntro} succeeds, it learns a new rule $\rho_d$
in one of two ways: either (1) by adding to the body of $\rho_f$ an assumption $\alpha(X)$ already belonging to $\A$
or (2) by adding to the body of $\rho_f$ a new $\alpha(X)$ and introducing its contrary $c\_\alpha(X)$.
In case (1) $\langle \RR \cup \RR_l \cup \{\rho_d\}, \A,\!\contrary \rangle$ is a solution, as guaranteed by the condition at line \ref{alg:a5}.
In case (2) $\Babalearn$ gets a solution after learning a suitable set of facts for $c\_\alpha(X)$ (lines \ref{alg:g12}--\ref{alg:g13}), as guaranteed by Proposition~\ref{prop:asm}.
Thus, in both case (1) and case (2), the invariant of the foreach-loop of \textit{Gen} is preserved, that is, $\langle \RR\cup\RR_l,\A,\!\contrary\rangle$ bravely entails $\EE$. 

If $\Babalearn$ terminates with success, then $\RR_l=\emptyset$ and $\langle \RR,\A,\!\contrary\rangle$ is a solution of $(\sABAo,\EE,\T)$.

\section*{Proof of Proposition \ref{prop:foldterm}}

Since $\LL$ is a finite language, condition $F2$ enforces that there exists a finite set $\mathcal B$ of atoms such that, for any $\rho_f$ obtained from $\rho$ by any sequence of applications of $\textit{fold}$, $bd(\rho_f)\subseteq \mathcal B$  (up to a renaming of variables).
Thus, repeated applications of $\textit{fold}$ will eventually output a rule $\rho_f$ such that $\textit{foldable}(\rho_f,\RR)$ is false and \textit{applyFolding}
will terminate. By condition $F1$, $\rho_f$ is intensional.

\section*{Proof of Theorem \ref{thm:completeness}}

\noindent
This proof is like the one for Theorem~\ref{thm:soundness}, with the additional property that, when applied within $\EBabalearn$, the \textit{Gen} procedure will always terminate with success.
Indeed, after having tried a number of applications of the Folding transformation, followed by a failure due to \textit{applyAsmIntro} (line \ref{alg:a4}), 
$\textit{applyFolding}(\rho, \RR)$ will return an unchanged $\rho$. 
Thus,  \textit{RoLe} computes a solution of the input brave ABA Learning problem $(\sABAo,\EE,\T)$, if any solution exists (by Theorem~\ref{thm:sol}), and 
\textit{Gen} terminates with a (possibly non-intensional) solution as output.

\end{document}

%% file: algorithm.tex
\newcommand\mycommfont[1]{\footnotesize\rmfamily{#1}}
\SetCommentSty{mycommfont}

\begin{algorithm}
\KwIn{$(\sABAo,\EE,\T)$: learning problem}
\KwOut{$\sABA$: intensional solution }
\SetKwProg{roLe}{Procedure}{}{}
\SetKwProg{gen}{Procedure}{}{}
\SetKwProg{applyFolding}{Function}{}{}
\SetKwProg{applyAsmIntro}{Function}{}{}

$\RR$ := $\RRin$; ~
$\A$ := $\A_0$; ~ 
$\contrary$ := $\contrary^{_0}$; ~
$\RRl$ := $\emptyset$\; 

$\textit{RoLe}()$; ~ $\textit{Gen}()$; ~ \Return{$\sABA$};

\roLe{RoLe$()$}{\label{alg:role}

 $P$ := $\mathit{ASP}(\langle\RR,\A,\contrary\rangle,\EE,\T)$\;\label{alg:r1}

 \eIf{$\neg\textit{sat}(P)$}{
   fail\;\label{alg:r2}
 }{
 $S$ := \textit{getAS}($P$)\;\label{alg:r3}
 \tcc{- ~Rote learning~ - - - - - - - - - - - - - - - - - - - - - - - - - - - - - - }
 \ForEach{$\mathtt{p'(t)} \in S$}{\label{alg:r4} 
   $\RRl$ := $\RRl\cup\{p(X) \leftarrow X\!=\!t\}$\;\label{alg:r5} 
  }\label{alg:r6} 
 }
}\label{alg:endrole}
\gen{\textit{Gen$()$}}
{
\ForEach{$\rho: (p(X) \leftarrow X=t) \in \RRl$}{\label{alg:g1}

  $\RRl\!:=\!\RRl\!\setminus\!\{\rho\}$;\quad 
  $\RRt\!:=\RR\cup\RRl$;

  \tcc{- ~Fact subsumption~ - - - - - - - - - - - - - - - - - - - - - - - - - - - -}

  \If{$\neg\,\textit{sat}(\mathit{ASP}(\langle\RRt,\A,\!\contrary\rangle,\EE,\emptyset))$}{\label{alg:g14}
  
    \tcc{- ~Folding~ - - - - - - - - - - - - - - - - - - - - - - - - - - - - - - - -}
  
    $\rho_f$\,:=\,\textit{applyFolding}($\rho,\RR$);\label{alg:g2} \vspace{1pt}


  \If{$\neg\,\textit{sat}(
  \mathit{ASP}(\langle 
    \RRt \cup \{\rho_f\}, \A,\!\contrary \rangle,
  \langle\Ep\!,\En\rangle,\emptyset))$}{\label{alg:g3}
  
  \tcc{- ~Assumption introduction~ - - - - - - - - - - - - - - - - -}
  
  
  $\langle\rho_d,\alpha(X),S\rangle$ := \textit{applyAsmIntro}($\rho_f,\RRt$);\label{alg:g5}

  \tcc{
  \hspace*{1pt}(1)~$\rho_d$ is a rule of the form
      $~H\!\leftarrow\!B,\alpha(X)$\break
  \hspace*{1pt}(2)~$\alpha(X)$ is an assumption with $X\!\!=\!\textit{vars}(B)$\break
  \hspace*{1pt}(3)~$S$ is a set of atoms including also those for \break \hspace*{11pt} the contrary 
  $c\_\alpha(X)$ of $\alpha(X)$}
  
  $\RR$\,:=\,$\RR\!\cup\!\{\rho_d\}$;
  
  $\A$\,:=\,$\A\!\cup\!\{\alpha(X)\}$;
  
  $\overline{\alpha(X)}$\,:=\,$c\_\alpha(X)$\;\label{alg:g6} 
  
  \tcc{- ~Rote learning~ - - - - - - - - - - - - - - - - - - - - - - - - -}

  \ForEach{$\mathtt{c\_\alpha(t)} \in S$}{\label{alg:g12}
   $\RRl$ := $\RRl \cup \{ {c\_\alpha(X) \leftarrow X\!=\!t}\}$\;
  } \label{alg:g13}
  }
  }
} 

} 
\applyFolding{$\textit{applyFolding}(\rho,\RR)$}{
 
 \While{$\textit{foldable}(\rho,\RR)$}{\label{alg:g17}
 
   $\rho := \textit{fold}(\rho,\RR)$;
  
 }
 return $\rho$\;
}
\applyAsmIntro{applyAsmIntro($H\leftarrow B,\RR$)}{\label{alg:asm}

  $X$ := $\textit{vars}(B)$\;\label{alg:a1} 

  \eIf{there exists $\alpha(X)\in\A$ relative to $B$ }{\label{alg:a2} 
    $\rho$\,:=\,$H\!\leftarrow\!B, \alpha(X)$; \quad $S$\,:=\,$\emptyset$
    
    \If{$\neg\textit{sat}(ASP(\langle\RR\cup\{\rho\},\A,\contrary\rangle,\EE,\emptyset))$}{\label{alg:a4}
      fail\;\label{alg:a5} 
    }
  }(\tcc*[h]{introduce an assumption $\alpha(X)$, with a new predicate $\alpha$}){\label{alg:a6}  
    $\rho\,$ := $H\!\leftarrow\!B, \alpha(X)$\;
    
    $F$ := $\langle\RR\cup\{\rho\}\!,\A\cup\{\alpha(X)\}\!,\contrary\cup\{\alpha(X)\!\mapsto\!c\_\alpha(X)\}\rangle$\;
    
    $S$ := $\textit{getAS}(\mathit{ASP}(F,\EE,\{c\_\alpha\}))$\; \label{alg:a9} 
  }
  return $\langle \rho,\alpha(X),S\rangle$\;\label{alg:a10} 
  
}

\caption{$\Babalearn$\label{alg:strategy}}
\end{algorithm}